\documentclass[letterpaper, 10 pt, conference]{ieeeconf}  

\IEEEoverridecommandlockouts                              

\usepackage{listings}
\usepackage{amssymb}
\usepackage{caption}
\usepackage{multirow}
\usepackage{wrapfig} 
\usepackage{graphicx} 
\usepackage{lipsum} 
\usepackage[product-units=repeat]{siunitx}
 \usepackage{xcolor}
\usepackage{graphicx}
\usepackage{amsmath}
\usepackage{cite}
\usepackage{subfig}
\usepackage{hyperref}

\title{\LARGE \bf

From Human Hands to Robotic Limbs: A Study in Motor Skill Embodiment for Telemanipulation 
}

\author{Haoyi~Shi$^{1}$, Mingxi~Su$^{1}$, Ted~Morris$^{1}$,  Vassilios~Morellas$^{1}$, and Nikolaos~Papanikolopoulos$^{1}$%
\thanks{$^{1}$Haoyi~Shi, Mingxi~Su, Ted~Morris, Vassilios~Morellas, and Nikolaos~Papanikolopoulos are with the Minnesota Robotics Institute (MNRI), University of Minnesota
{\tt\footnotesize $\{$shi00317 $|$  su000111 $|$ tmorris $|$ morellas $|$ papan001$\}$@umn.edu}}%
\thanks{\url{https://sites.google.com/umn.edu/teleop/home}}%
}

\begin{document}

\maketitle
\thispagestyle{empty}
\pagestyle{empty}
\begin{abstract}

This paper presents a teleoperation system for controlling a redundant degree-of-freedom (DOF) robot manipulator using human arm gestures. We propose a GRU-based Variational Autoencoder (VAE) to learn a latent representation of the manipulator’s configuration space, capturing its complex joint kinematics. A fully-connected neural network maps human arm configurations into this latent space, allowing the system to mimic and generate corresponding manipulator trajectories in real-time through the VAE decoder. The proposed method shows promising results in teleoperating the manipulator, enabling the generation of novel manipulator configurations from human gestures that were not present during training. Experimental evaluation highlights the effectiveness of this approach, although limitations remain concerning training data diversity.

\end{abstract}

\vspace{-4.5mm}
\section{Introduction}
The convergence of Artificial Intelligence(AI) and robotic systems, has revolutionized a wide range of domains such as agriculture, healthcare medicine, warehousing, and manufacturing. In recent years, generative AI models for image and language generation (Large Language Models, LLMs) have proliferated, such as for image generation and natural language generation, for example ChatGPT, Midjourney, and Dall-E, to name a few. The combination of generative models and robots has also started to receive considerable attention. One aspect that has impeded its progress is that generative models are data-driven approaches. At the onset of LLMs, and image generation, copious available images and text on the Internet were mined for building massive training datasets. Training generative models for robotic applications requires a substantial amount of domain- and task-specific data, which, on the contrary, does not exist~\cite{hamalainen_affordance_2019}. There is no protocol or standard for a taxonomy of robotic trajectory and task data for generating robotic actions for new operative scenarios. This is because the application scenarios are constrained. In theory, teleoperation technology can address these limitations by allowing the human operator to simultaneously supply new training exemplars to teach the robot to perform unforeseen tasks and to augment training sets for new task scenarios. 

Manipulator teleoperation systems enable remote interaction with environments and can scale human motion to achieve larger or smaller action capabilities. These systems aim to accurately translate human decision-making and actions while ensuring the robust operation of the teleoperation system~\cite{hirche_human-oriented_2012}. A common solution is for the operator to control the real-time end-effector's position and orientation with external devices with a $6$ degrees of freedom (DOF) robot with the joint trajectory space calculated through standard inverse kinematics (IK). For example, an operator can use a haptic device to control the end-effector 6-DOF representation to achieve high-frequency teleoperation for industrial manipulators in workplace and factory settings~\cite{dekker_design_2023}. Another approach instead uses an external RGB and RGBD (depth) camera to estimate the operator's $6$-DOF hand pose~\cite{schroder_real-time_2012,antotsiou_task-oriented_2018} for teleoperating the robot end-effector ~\cite{ajili_gesture_2017,qin_anyteleop_2024}. Furthermore, as the virtual reality (VR) hardware market expands, inertial measurement unit (IMU) sensors and VR controllers are becoming a commonly used approach to estimate the operator's $6$-DOF hand pose for integrating $6$-DOF robot teleoperation and control ~\cite{lipton_baxters_2018,weigend_anytime_2023}.
\begin{figure}[!t]
    \centering
    \includegraphics[width=0.75\linewidth]{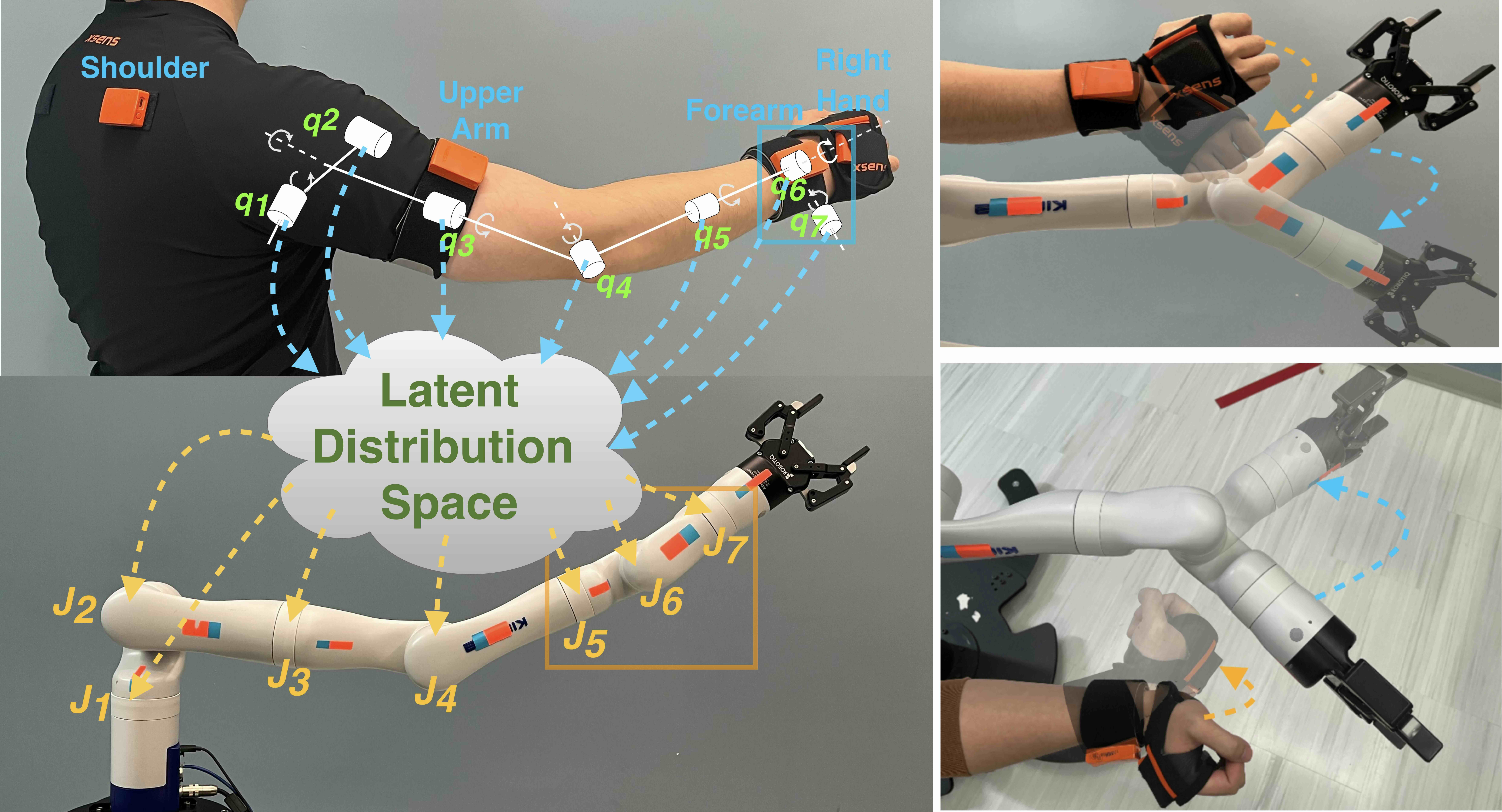}
    \captionsetup{font=footnotesize}
    \caption{\textbf{Orange} labels, $\mathbf{\textit{J}_1\textit{-}\textit{J}_7}$, indicate the positions of the {K}inova manipulator's joints. \textbf{Green} labels, $\mathbf{\textit{q}_1\textit{-}\textit{q}_7}$, indicate the positions of the human arm's joints. \textbf{Arrow} shows the mapping relationship between the manipulator's joint and the kinematic chain of a human upper limb.}
    \label{fig:jointMapping}
    \vspace{-25pt}
\end{figure}
These approaches are less effective when applied to redundant DOF manipulators with more than six degrees of freedom. Redundant DOF manipulators afford multiple solutions in joint configuration space to achieve the same end-effector pose. Although this kinematic redundancy provides greater flexibility within complex and dynamic environments, it also increases the complexity of teleoperation. To address this problem and provide an intuitive human-robot interface, we propose a novel, computationally efficient, machine learning-based approach to determine plausible robot joint and manipulator trajectories that mimic human kinematic motion behaviors, making it an ideal interface for co-robotics task scenarios. The approach embodies a training-model learning framework that can reduce or eliminate the typically highly technical and labor intensive requirement to re-program such robots to execute new tasks and operate in new environments. 

Specifically, we present a Gate Recurrent Unit based Variational Autoencoder (GRU-based VAE) architecture finding the latent representation of the redundant DOF robot manipulator configuration space. A feed-forward neural network converts human arm gestures into the latent distribution space and generates the corresponding robot manipulator configuration trajectory by trained VAE decoder in real-time. In our approach, the resulting human-robot action coupling during tasks can be used to further create new imitation learning samples. The simulation based model pipeline created to train the model can also be easily adapted to train the model for other robot configurations. 











\vspace{-3.5mm}

\section{related work}
\vspace{-2.5mm}
One approach to achieve high-DoF manipulator teleoperation is Master-Slave or Twin-Master, where the operator manually controls another manipulator with identical kinematics to the target manipulator~\cite{singh_haptic-guided_2020,su_heterogeneous_2021}. However, this method requires an additional manipulator with the same kinematic structure as the remotely controlled device. This requires either two identical manipulators or building a custom one. Both options are costly and limit the manipulator system's adaptability to other types of manipulators. 

A Human arm is defined as a $7$-DOF kinematic structure~\cite{prokopenko_assessment_2001}. Hence, teleoperating a high-DOF manipulator by mapping the human arm joints to the target manipulator is another approach. Previous work~\cite{penco_robust_2018}, achieved humanoid robot ($7$-DOF arm) teleoperation by attaching IMU sensors to the human operator and mapping whole-body joints with dynamic filters between the operator and robot. However, a limitation of this method is that the required robot kinematic structure must be similar to the human arm kinematic structure. Other strategies have been explored to simplify the kinematic representation of the human arm for teleoperation~\cite{ajoudani_reduced-complexity_2018,su_deep_2019,arduengo_human_2021}. For example,~\cite{su_deep_2019} introduced elbow elevation angle as a constraint for a human arm mapped on the robot as a swivel motion. 
~\cite{arduengo_human_2021} proposes a method that separates a redundant $7$-DOF manipulator as a $3$-DOF manipulator attached to a $4$-DOF end-effector. Using IMU sensors to couple the operator's hand with $3$-DOF end-effector and elbow with $4$-DOF end-effector position changes, with IK calculation for both sub-manipulators' joints in real-time to achieve teleoperation.

Our work draws inspiration from the concept of modifying kinematic representations. While most manipulators use a combination of single DOF revolute or prismatic joints, human arm articulation kinematics are represented by a combination of ball-and-socket and condyloid joints to achieve complex joint movements. Finding robot combinatorial joint kinematics that can mimic complex human articulated arm motion is an essential step in our method. Specifically, and instead of imposing explicit constraints to redundant DOF robots, we propose a generative deep learning (DL) framework to find a low-dimensional robust implicit, kinematic representation that describes complex redundant DOF robotic motion. Our framework devises a GRU-based variational autoencoder deep neural network (DNN) that generates robot trajectories which faithfully mimic human gestural intent for completing tasks.

A Variational Autoencoder (VAE) is a neural network architecture with an encoder, a latent space, and decoder module\cite{kingma_auto-encoding_2022}. It was first introduced as an image-generative model that efficiently uses a neural network to approximate the likelihood function for the latent space distribution, which is modeled as a mixture of multiple Gaussian distributions derived from the training dataset. By sampling latent space features from Gaussian distributions, we can generate a new, unforeseen image by the decoder, which endows the characteristic features of an actual image. Such an architecture can also be used to prescribe $3$D physical motions. ~\cite{hamalainen_affordance_2019} introduced a method that integrates two Variational Autoencoder (VAE) modules: one for processing visual sensor data and the other for the manipulator's trajectory. By utilizing the latent representation from the vision module, the decoder of trajectory-VAE can generate movement paths for object-reaching tasks in the $2$D Cartesian plane.

On the other hand, the Recurrent Neural Networks (RNNs) are the most commonly used methods for sequential data prediction and feature extraction. For example, \cite{weigend_anytime_2023} utilizes Long Short-Term Memory (LSTM) as a neural network module to predict the human elbows and wrists position in order to implement an intuitive control for manipulator based on the time series data from an IMU sensor in a smartwatch.

 VAEs and RNNs can be integrated together for inference tasks. For example \cite{bowman_generating_2016}, proposed an LSTM-based VAE for missing word-imputing tasks. The LSTM allows the VAE model to consider global concepts of a sentence and generate more diverse and well-formed sentences compared with the standard RNN language model. Furthermore, in \cite{ozdemir_embodied_2021}, they also present an LSTM-based VAE framework for robot-embodied Language Learning. This allows the system to generate the correct action description by executing the action. 

The remainder of the paper is as follows. Section $3$ discusses the proposed teleoperation system, including the training dataset, the proposed GRU-based VAE model for discovering the manipulator's latent distribution space, and the fully-connected neural network module used to map human arm configurations to the manipulator's latent space, along with the model training results. Section $4$ outlines the experimental design and presents the results used to evaluate the proposed teleoperation system. Section $5$ addresses the limitations of the current method, while Section $6$ provides the conclusion. This study is under the IRB: STUDY$00009131$.


\vspace{-3.5mm}
\section{methodology}
\label{sec:methodology}
\vspace{-2mm}


In this work, we propose a novel method that utilizes a VAE neural network architecture to learn a latent distribution space that can represent complex joint movements for a 7-DOF Kinova manipulator. By leveraging this probabilistic distribution, we can approximate the Kinova configuration space with limited training data. We also train a feed-forward neural network to map human arm kinematics to the learned latent space and use the VAE decoder to generate corresponding manipulator joint configurations. This approach involves creating datasets for both Kinova trajectories and human arm configurations. The following section details the data collection, model architecture, and overall workflow.

\vspace{-2.2mm}

\subsection{Data Collection}
\vspace{-1mm}
\subsubsection{\textbf{\textsc{Kinova trajectory dataset collection}}}\label{kinova_data}

We randomly select $500$ start and end \textit{x}, \textit{y}, \textit{z} end-effector positions in Cartesian space, and assign two corresponding random sets of valid Roll($\phi$), Pitch($\theta$), Yaw($\psi$) orientation for each start and end position, i.e.,  $\mathbf{p} = [\textit{x}, \textit{y}, \textit{z}, \phi, \theta,\psi]^\top, \mathbf{p} \in \mathbb{R}^{6}$. Inspired by \cite{hamalainen_affordance_2019}, the MoveIt software is used to generate corresponding trajectories, including joint angles under varied time steps for every combination of initial start and end poses. Note that a jump from $-180$ degrees to $179$ creates a discontinuity singularity of the collected joint angular data that can confound training the model ~\cite{weigend_anytime_2023}. 
To mitigate such singularities, each manipulator's angle $\mathbf{\textit{d}}$ is converted to projected unit values $\mathbf{(\textit{cos(d)}, \textit{sin(d)})}$, which therefore results in $14$ values to represent the $7$-DOF manipulator. We use the Cubic Spline interpolation algorithm to appoint each trajectory time step into $0.1$s ($10$Hz) intervals. Finally, we decompose each trajectory as a sequence of $2$ time-step segments that represent the current and the next time-step trajectory features.

\subsubsection{\textbf{\textsc{Kinematic mapping}}}
To achieve intuitive and efficient manipulator teleoperation with the human arm, a kinematic correlation between the manipulator and the human arm was first defined before the model training and data collection.  We collect the operator's right-arm-related data for training. As shown in Fig.~\ref{fig:jointMapping}, $\mathbf{\textit{J}_1,_2}$ are mapped to the shoulder joint $\mathbf{\textit{q}_1,_2}$, $\mathbf{\textit{J}_3}$ is mapped to the upper arm joint $\mathbf{\textit{q}_3}$, $\mathbf{\textit{J}_4}$ is mapped to the elbow $\mathbf{\textit{q}_4}$, and $\mathbf{\textit{J}_5}$ is mapped to the joint rotation of the forearm arm. However, $\mathbf{\textit{J}_5,_6,_7}$ is also considered a universal joint representing the human wrist's spherical joint. Therefore, Joint $\mathbf{\textit{J}_5}$ overlaps between the representation of forearm arm and wrist rotation. When $\mathbf{\textit{J}_5,_7}$ represents a flexion or extension action of the wrist, $\mathbf{\textit{J}_5}$ should rotate conditionally to allow the direction of the rotation axis of $\mathbf{\textit{J}_6}$ to match the rotation axis ($\mathbf{\textit{q}_7}$) of the wrist's flexion or extension action, and $\mathbf{\textit{J}_7}$ will rotate oppositely to counteract the unwanted rotation from $\mathbf{\textit{J}_5}$ to keep the facing direction for the back of human hand and the end-effector consistent. On the other hand, if $\mathbf{\textit{J}_5,_7}$ represents either an ulnar or radial deviation ($\mathbf{\textit{q}_6}$), $\mathbf{\textit{J}_5}$ will only repose to the forearm arm rotation ($\mathbf{\textit{q}_5}$), $\mathbf{\textit{J}_7}$ should not rotate. In the end, $\mathbf{\textit{J}_7}$ is not mapped to any human joint but is only used to counteract the rotation from $\mathbf{\textit{J}_5}$. 

\subsubsection{\textbf{\textsc{Human arm joints configuration dataset collection}}}\label{awinda_data}
Once the VAE model is settled, we can construct the human arm joint configuration dataset using the XSens Awinda human skeletal motion tracking system. Wireless $6$-DOF IMU sensors were attached to 11 upper human body segments (sternum, pelvis, head, L/R hands, L/R forearms, L/R upper arms, and L/R scapular skeletons joints) to approximate complete upper body skeletal motion. Table~\ref{tab:data-table} shows the details for the data of joints collected for the manipulator and the participant's right arm, including the joint's name for both and the joint's range only for part one data of the manipulator. An upper body calibration procedure using proprietary software was required to allow the IMU sensors to correctly track the full upper-body human kinematic motion~\cite{schepers_xsens_2018}. Although optoelectronic target point rigid body tracking systems have been typically used for Telerobotics research due to their high precision and accuracy, the inertial measurement sensor based systems can be deployed in a myriad of more diverse environments containing complex spatial constraints and occlusions. The premise of our approach, which is thus far supported by experiments described herein, is that the measured human kinematics only require to be \emph{repeatable} since the human can adjust their visuo-proprioceptive queues according to context and nature of the task to be completed. 

\begin{table}[h!]
\scriptsize 
\centering
\begin{tabular}{|c|c|l|}
\hline
\textbf{Categories} & \multicolumn{2}{c|}{\textbf{Collected Features}} \\ \hline
\textbf{\begin{tabular}[c]{@{}c@{}}Kinova Joints\\ Angle (Degree)\end{tabular}} &
  \multicolumn{2}{l|}{\begin{tabular}[c]{@{}l@{}}Joint 1 (-15°$\sim$90°), Joint 2 (60°$\sim$120°), \\ Joint 3 (-15°$\sim$90°), Joint 4 (-120°$\sim$-60°), \\ Joint 5 (-90°$\sim$180°), Joint 6 (-90°$\sim$30°), \\ Joint 7 (0°$\sim$-90°)\end{tabular}} \\ \hline
\multirow{2}{*}{\textbf{\begin{tabular}[c]{@{}c@{}}Awinda Joints\\ Angle (Degree)\end{tabular}}} & \textbf{\begin{tabular}[c]{@{}c@{}}T4-Shoulder,\\ Shoulder\end{tabular}} & \begin{tabular}[c]{@{}l@{}}1. Abduction/Adduction \\ 2. Internal/External Rotation \\ 3. Flexion/Extension\end{tabular} \\ \cline{2-3} 
 & \textbf{Elbow, Wrist} & \begin{tabular}[c]{@{}l@{}}1. Ulnar Deviation/Radial \\ Deviation \\ 2. Pronation/Supination \\ 3. Flexion/Extension\end{tabular} \\ \hline
\end{tabular}
\captionsetup{font=footnotesize}
\caption{The data collection participant is strongly right-handed. Hence, only right-arm data features are used for training.}
\label{tab:data-table}
\vspace{-5pt}
\vspace{-1.5mm}
\end{table}

The human arm joint configuration dataset has a total of $86$ actions and consists of two parts. The first part contains $74$ actions, in which only a single joint participates in each action from the operator's perspective, for example, shoulder abduction or adduction, wrist pronation or supination, and so on. For the second part, multiple joint movements are used in each action. From a randomly sampled large group of start and end poses, only a small subset is feasible to physically mimic (e.g., imitate) the human operator's right arm. For the initial training of the DNN, a feasible training set of $12$ robot start and end poses and the corresponding human operator arm trajectories were collected. Then, we execute each action in a real manipulator to collect real-time joint angle data. The collected raw joint angle data from the manipulator and the human joint angle data, which will be described below, are both then converted to projected unit values as described in~\ref{kinova_data}. Then, the processed angle dataset is put into the VAE encoder to generate a sequence of latent representations of each action. In parallel, the human operator observes the kinematic motion of the robot manipulator and attempts to mimic the motion with corresponding joints as described in Fig.~\ref{fig:jointMapping}. This process facilitates proprioceptive neuromuscular adaptation, commonly referred to as 'muscle memory'. Once the operator feels adequately familiar with the demonstrated actions, they proceed to use their right arm to mimic each movement, recording the entire process to capture the joint angles of the operator's arm. It is important to note that the absolute positions of the manipulator's end-effector and the human hand may differ due to the disparity in arm lengths. Since the action completion time and the data reception rate between the manipulator and human arm cannot be synchronized, the manipulator data was temporarily matched to the closest time steps of the human operator arm joint data through the Cubic Spline interpolation algorithm and then placed together as a time synchronized pair. In total, we created $15,043$  pairs of human gestural arm configurations and corresponding latent representations of robot manipulator configurations with a selected frequency of 40 Hz (0.025 second intervals). Note that only one participant was involved in the human data collection process. The participant is $170$ \textit{cm} tall and of average body shape.
\vspace{-2.2mm}
\subsection{Generative Models}
 \begin{figure}[t]
    \centering
    \includegraphics[width=0.9\linewidth]{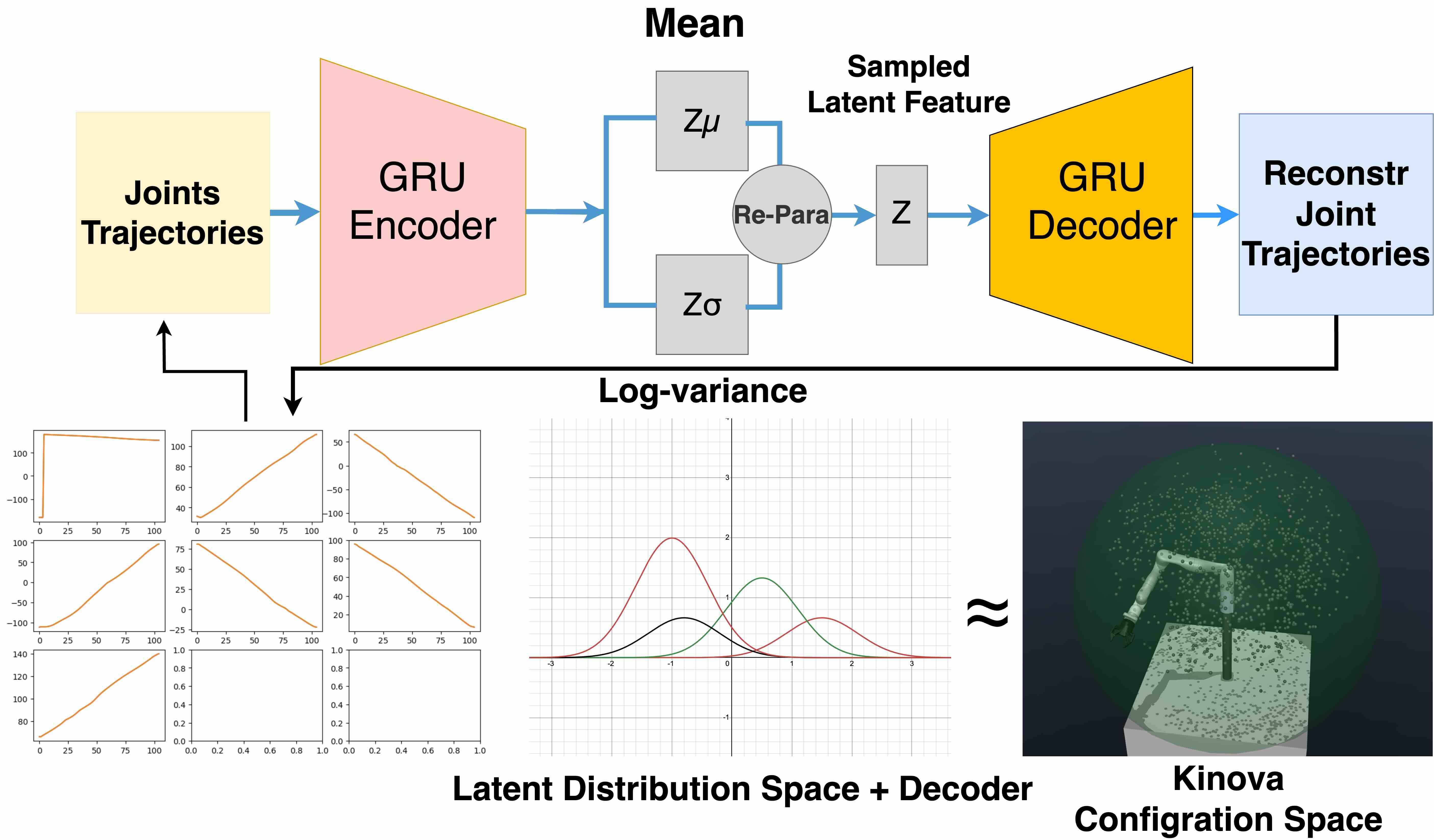}
    \captionsetup{font=footnotesize}
    \caption{The GRU-based VAE takes a 2-time-step manipulator joint angle position trajectory as input. It learns an approximate latent distribution by sampling latent features using the reparameterization trick, which is then passed to the decoder to reconstruct the input trajectory. The learned latent distribution space enables the approximation of the entire manipulator configuration space.}
    \label{fig:vae}
    \vspace{-20pt}
\end{figure}
\vspace{-2.2mm}

\subsubsection{\textbf{\textsc{GRU-based Variational Autoencoder}}} We utilize a latent variable model, specifically a Variational Autoencoder (VAE), to approximate and predict the configuration space of the manipulator within the continuous probability latent space. We use $2$ time-steps of joint position trajectory as the input to the VAE. The first time-step represents the manipulator's current joint positions, while the second time-step represents the next joint positions, $100$ milliseconds later. As discussed in the related work, the encoder and decoder are crucial for approximating a latent distribution $\mathcal{Q}({z}\mid{x})\text{ and likelihood distribution }\mathcal{P}({x}\mid{z})$.
 To effectively capture the time-sequence information and enhance prediction accuracy, we utilized the GRU recurrent-based neural network architecture, for both the encoder and decoder in the VAE, as shown in Fig.~\ref{fig:vae}. The encoder takes the Kinova robot trajectory as input feed into the GRU layer. Then, the selected features are passed to two separate single-layer neural networks that generate the mean $\vec{\mu}$ and log-variance $\vec{\sigma}$ to represent a finite number $z$ of Gaussian distributions to approximate the true latent space distribution, where the size of $z$ is a hyper-parameter. With the reparameterization trick equation~\ref{reparameterization} where noise \( \epsilon \sim \mathcal{N}(0, I) \), we can sample the latent feature $\vec{Z}$ for data reconstruction with the decoder and calculating a gradient and optimize the distribution when doing backpropagation during training. 
\vspace{-1.9mm} 
 \begin{equation}\label{reparameterization}
\vec{Z} = \vec{\mu} + \vec{\sigma} \cdot \vec{\epsilon}
\vspace{-15pt}
\end{equation}
\vspace{-1.9mm} 

To reconstruct the trajectory using the latent feature $\vec{Z}$ with the decoder,  $\vec{Z}$ must be repeated to match the length of the target trajectory. Specifically, the input to the decoder is a matrix where $\vec{Z}$ is repeated twice.

The loss function for the VAE model has two components: (1) the trajectory reconstruction loss and (2) the KL-divergence, which are represented by~\ref{vae_loss}.
\vspace{-6pt}
\begin{equation}\label{vae_loss}
    \mathcal{L}_{\text{VAE}} = \underbrace{\scriptsize\text{MAE}(\text{Input}, \text{Reconstr})}_{\text{Reconstruction Loss}} + \underbrace{\beta \cdot \scriptsize\text{KL}\left( q(z|x) \parallel p(z) \right)}_{\text{KL Divergence}}
\end{equation}
\vspace{-1.5mm}
\begin{equation}\label{kl_loss}
\small\text{KLD} = -0.5 \times \sum \left(1 + \vec{\sigma} - \vec{\mu}^2 - \exp(\vec{\sigma})) \right)
\vspace{-4pt}
\end{equation}
We choose the Mean Absolute Error(MAE) as the trajectory reconstruction loss to penalize the difference between the input and reconstructed trajectories.

The Kullback-Leibler (KL) divergence measures the deviation between latent distribution space and a Gaussian distribution, thereby enforcing each feature in the latent space to approximate a Gaussian distribution, \( \mathcal{N}(0, I) \), as shown in equation ~\ref{kl_loss} This regularization helps maintain the latent space's structure and facilitates downstream tasks.

The parameter $\beta$ helps control the percentage between the KL-divergence and reconstruction loss in the total loss. We need to optimize the decoder reconstructed joint trajectory performance as well as the learned approximation of the latent distribution space to the true likelihood distribution. A Sigmoid Annealing Schedule to $\beta$, then prevents the KLD vanishing problem during training~\cite{fu_cyclical_2019} which can reduce the interdependencies between each latent feature and specific robot joints during the training. 
\vspace{-12pt}

\begin{figure}[h]
    \centering
    \includegraphics[width=0.6\linewidth]{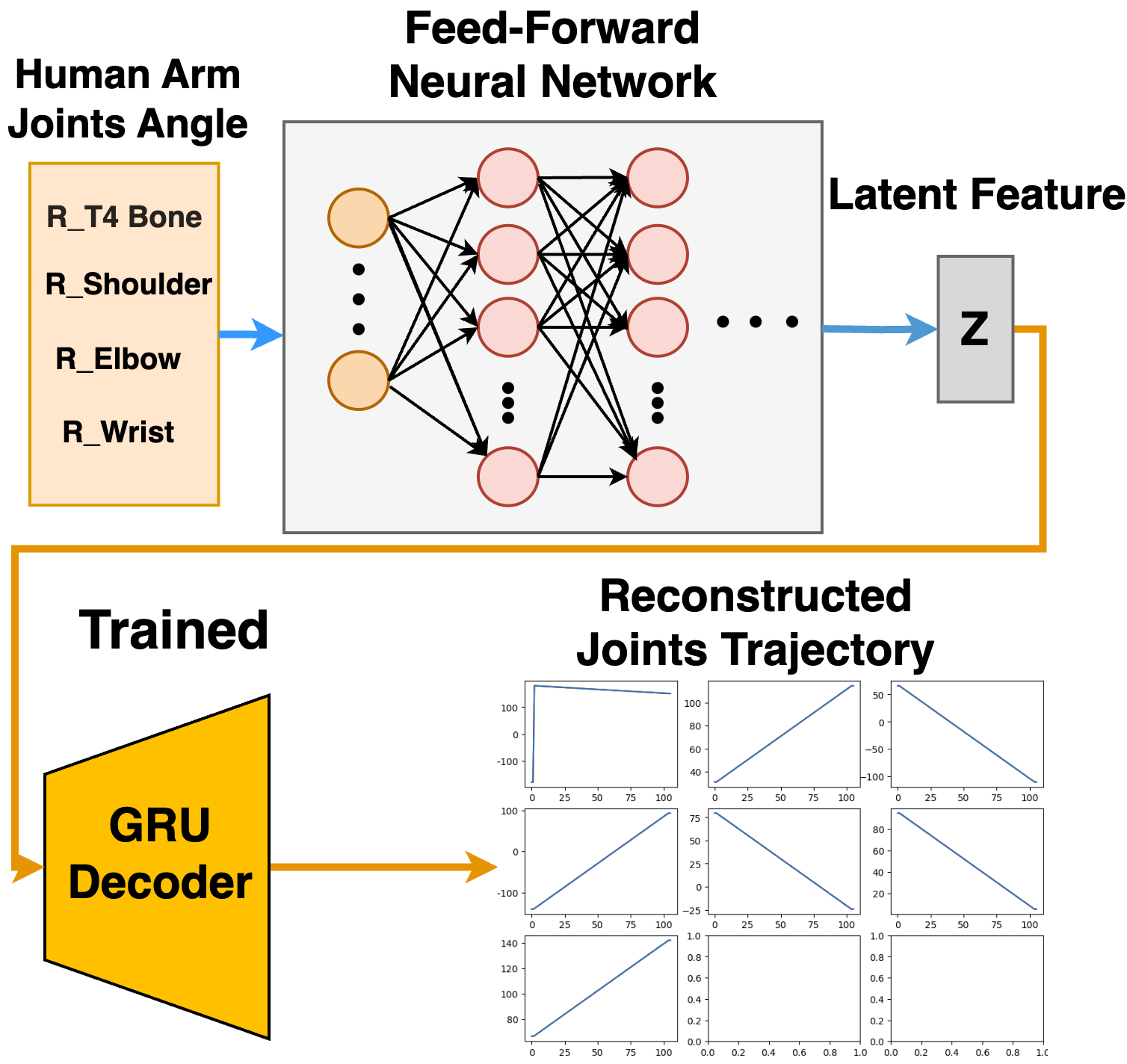}
    \captionsetup{font=footnotesize}
    \caption{The fully-connected module consists of an input layer, and three hidden layers with neuron counts ($32$, $40$, $20$). The output size is $10$ (latent feature size) for final predictions.}
    \label{fig:feedforward}
\end{figure}

\vspace{-4mm}

\subsubsection{\textbf{\textsc{Feed-forward neural network}}} To transfer the human arm joint configuration data to the executable configuration of the manipulator, we integrated a fully-connected neural network module shown in the top part of Fig.~\ref{fig:feedforward}. The fully connected module learns a data transformation from the human operator arm joints angle space to VAE latent feature space. We choose the Mean Absolute Error (MAE) as the loss function to penalize the error between the latent feature transformed from operator joints angle and target latent features, shown in \ref{mlp_loss}.
\vspace{-7pt}
\begin{equation}\label{mlp_loss}
\vspace{-6pt}
\mathcal{L}_{\text{mlp}} = \small\text{MAE}(\mathbf{y}, \hat{\mathbf{y}})
\end{equation}

In addition, we select the Scaled Exponential Linear Units (SELU) as the activation function for our fully-connected module. SELU's self-normalizing properties can help lower the difficulty of the learning process since each target latent feature follows a Gaussian distribution. The activation function is shown in equation~\ref{selu} where $\lambda \approx 1.0507$ \text{ and } $\alpha \approx 1.67326$.
\vspace{-8pt}
\begin{equation}\label{selu}
    \small\text{SELU}(x) = \lambda \begin{cases} 
    x & \text{if } x > 0 \\
    \alpha e^x - \alpha & \text{if } x \leq 0 
    \end{cases}
    \vspace{-5pt}
\end{equation}

The complete proposed teleoperation pipeline is illustrated in Fig.~\ref{fig:feedforward}. This pipeline integrates the trained GRU-based VAE decoder with a fully-connected neural network module. The decoder processes the latent features, which are passed through the fully-connected layers with the operator's right arm joints configuration to generate the desired outputs to control the robot's joint configurations.

\subsubsection{\textbf{\textsc{Model Training}}}

In the GRU-based VAE model, both the encoder and decoder use two GRU layers to extract features from $2$ time-step Kinova Gen3 7-DOF joint angle trajectories, with data shaped as $[2, 14]$ due to unit value conversion. After conducting multiple experiments, we set the hidden feature size to $28$ and the latent feature size to $10$. The annealing parameter $\beta$ reaches a maximum of $0.1$ and follows $4$ sigmoid cycles during training. The model was trained for $1,500$ epochs on an NVIDIA A$40$ GPU, with a batch size of $1,024,000$ and a learning rate of $1e-4$ using the Adam optimizer, taking approximately 5 hours.

To evaluate the performance of the GRU-based VAE, we first randomly sampled two latent feature vectors from the Gaussian distribution. Since both vectors represent different configurations of the Kinova robot, we use the VAE decoder to reconstruct/generate the corresponding configurations. One latent vector is designated as the starting point and the other as the endpoint. By interpolating $100$ steps between these two latent vectors and decoding the intermediate steps, we generate a joint angle trajectory list that captures the relationship between the latent features and the robot's joint angles. 

\begin{figure}[h]
    \centering
    \includegraphics[width=0.73\linewidth]{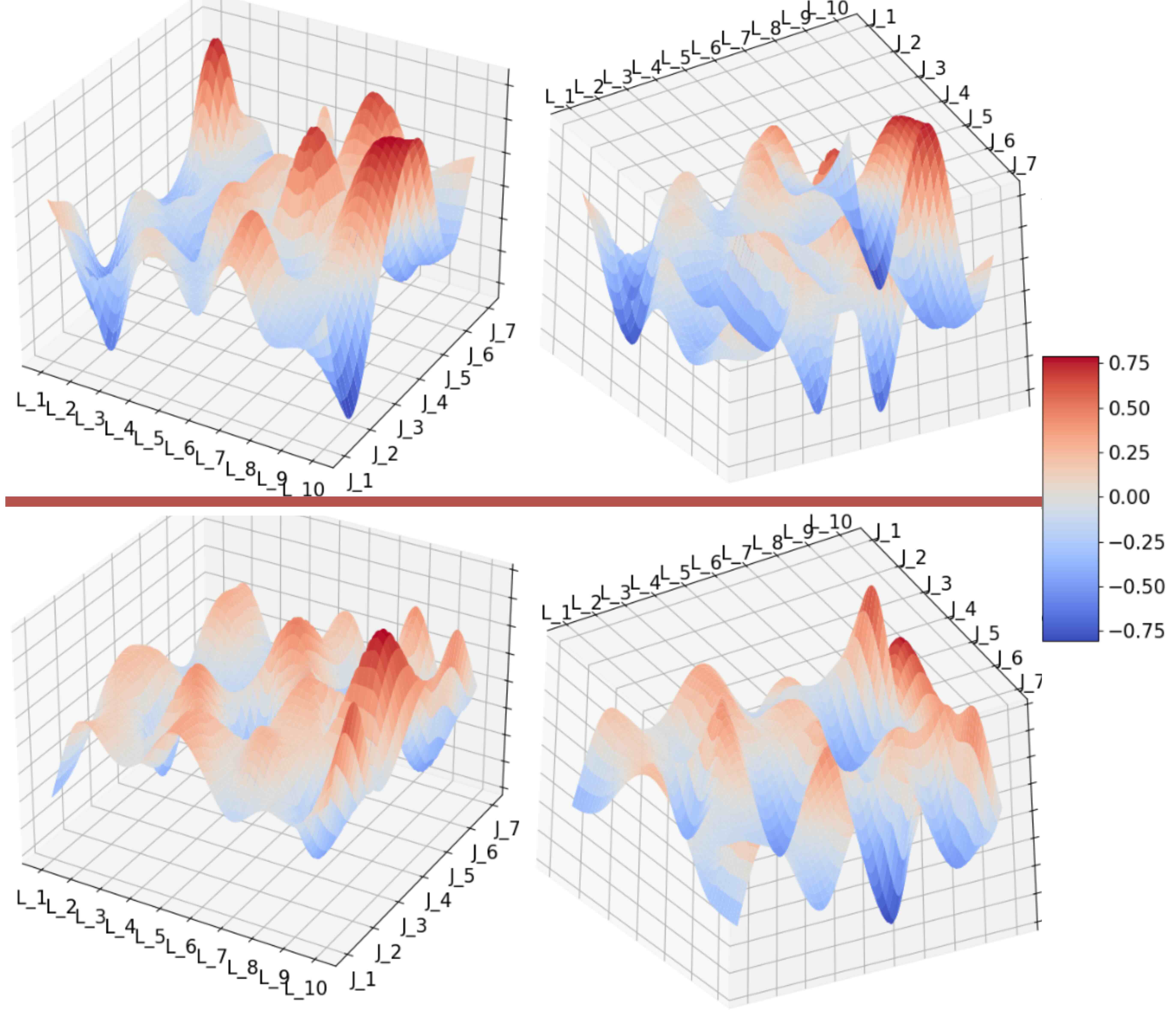}
    \captionsetup{font=footnotesize}
    \caption{This figure demonstrates the training result of GRU-based VAE with (Top) and without (Bottom) annealing scheduler. The \(Z\) axis represents the Correlation-Coefficient score for each Latent feature $L_1$ to $L_{10}$ and Kinova robot joint $J_1$ to $J_7$.}
    \label{fig:correlation}
    \vspace{-4.5mm}

\end{figure}

To comprehensively assess how each latent feature influences joint angles, we fix the first vector and repeat $10,000$ times for the rest of the pipeline. This approach provides detailed insights into the control each latent feature exerts over the joint angles, as shown in Fig.~\ref{fig:correlation}. For comparison, we trained two VAE models. The top results display the model training with the Sigmoid annealing scheduler, where specific latent feature displays a strong positive or negative correlation value with specific joint angles while having low correlation scores with other joints. Without the scheduler, the impact was more evenly distributed across latent features, making it harder for the fully-connected module to map human joint angles to manipulator features (involving one-to-many mapping; one \(X\) to many \(Y\)).

The fully-connected module consists of four feed-forward layers, with one dropout layers (rate of $0.5$) which are applied to the second layers to reduce over-fitting. The model is trained for $500$ epochs on an NVIDIA RTX $3060$, taking two minutes using the Adam optimizer with a learning rate of $1e-3$ and a batch size of $256$. The input size is $24$, as the $12$ joint angles are converted to projected unit values, and the output size is $10$, corresponding to the latent feature size of the GRU-based VAE model.
\begin{wrapfigure}{l}{0.18\textwidth} 
    \centering
    \includegraphics[width=0.2\textwidth]{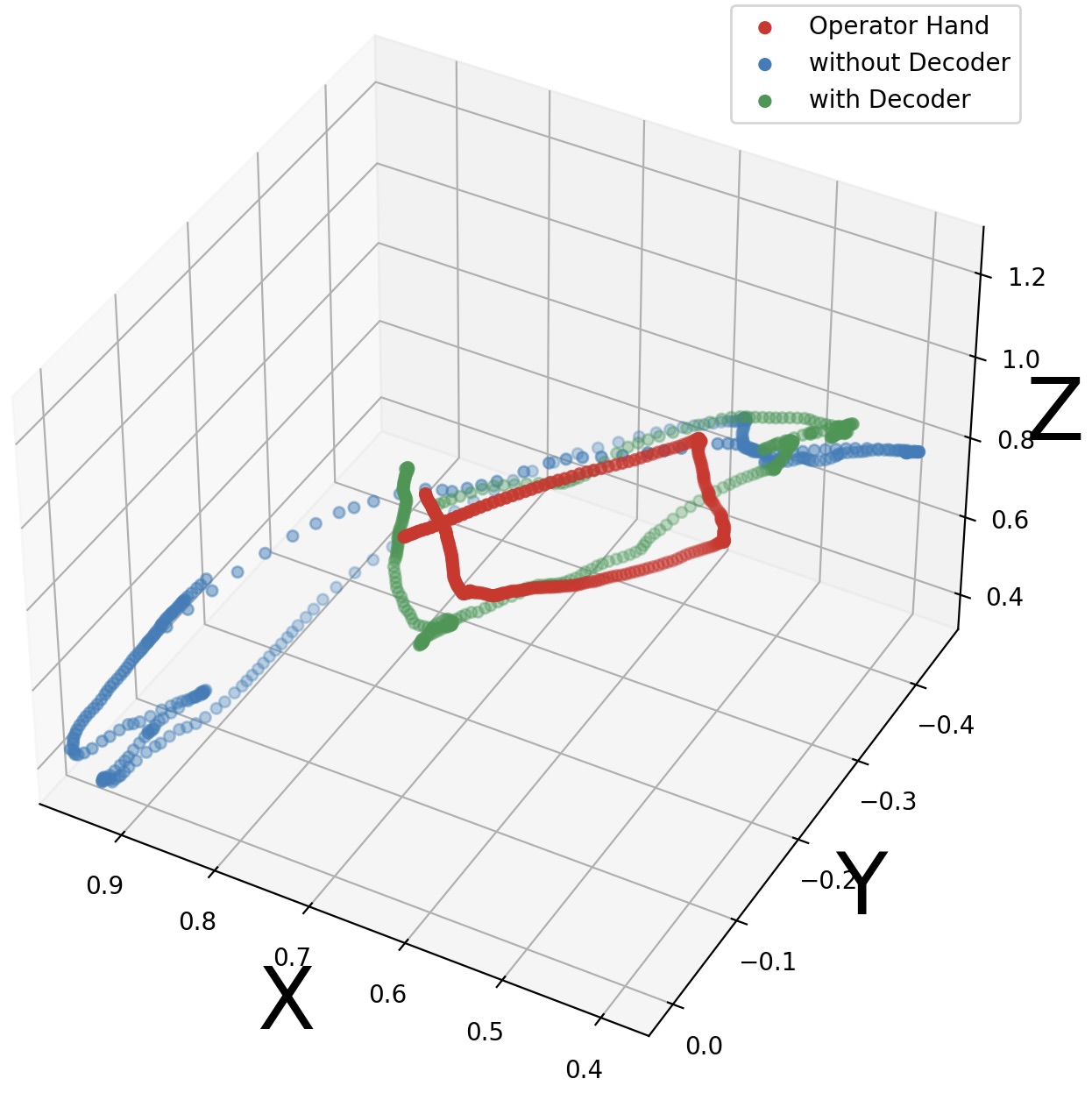} 
    \captionsetup{font=footnotesize}
    \caption{The \textbf{red} trajectory: operator's hand position in Cartesian space. The \textbf{green}: proposed teleoperation pipeline. The \textbf{blue}: fully-connected network without the VAE decoder.}
    \label{fig:comparsion_zx}
\vspace{-5pt}
\end{wrapfigure}

Theoretically, the VAE decoder approximates the likelihood distribution function \(\mathcal{P}({x}\mid{z})\) that enables it to generate new manipulator joint configurations. This approach has shown successful results in prior work, such as in image generation \cite{kingma_auto-encoding_2022} and in generating English sentences \cite{bowman_generating_2016}. By leveraging the continuity and smoothness of the probabilistic distribution, we can train the model with a limited set of paired human joint configurations and their corresponding latent representations of Kinova joint configurations. The VAE decoder is then able to interpolate and generate new Kinova joint configurations that correspond to human arm movements, even for configurations never encountered in the training process (uncertainty).


For comparison, we trained an alternative fully-connected neural network with the same architecture, except its output layer directly predicts $14$ Kinova joint angles. Using the same dataset of paired human and Kinova configurations, we pre-recorded a human arm trajectory and fed it into both models to generate corresponding manipulator trajectories. As shown in Fig~\ref{fig:comparsion_zx}, the proposed VAE-based pipeline outperforms the alternative in translating human arm movements and generating new manipulator configurations.
 

\vspace{-1mm}
\section{experiment and results}
\label{sec:experiments}

\begin{figure}[t]
    \centering
    \includegraphics[width=0.7\linewidth]{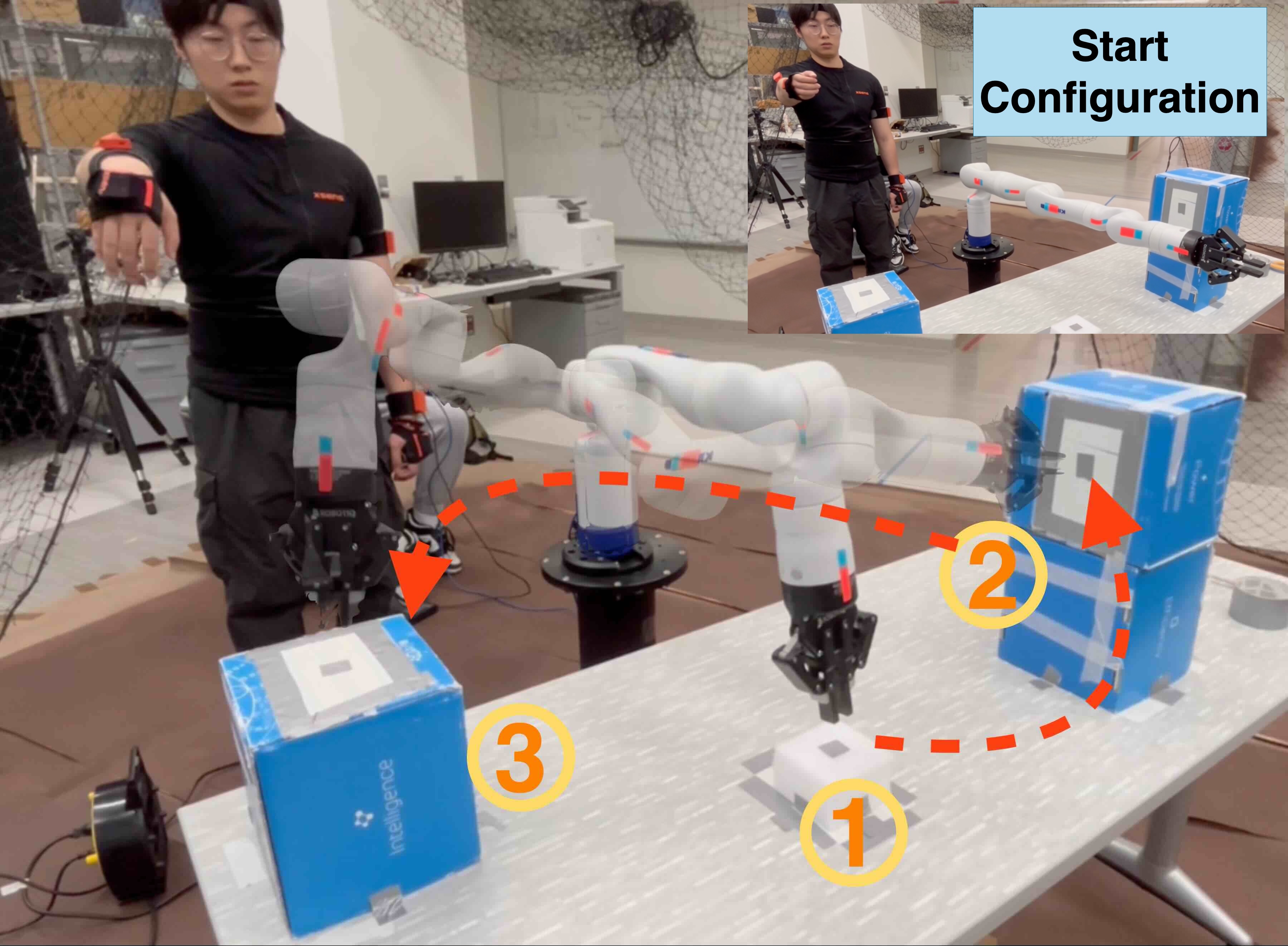}
    \captionsetup{font=footnotesize}
    \caption{Experiment setup. The operator needs to teleoperate the Kinova Gen$3$ $7$-DOF manipulator to reach three designated target poses. }
    \label{fig:experiment}
    \vspace{-7mm}
\end{figure}

The experimental setup is shown in Fig~\ref{fig:experiment}. It shows three predefined target poses and the corresponding target area, which is the gray bounding box, and the gray dot shows the center of the target area, providing visual information for participants. Each participant is required to teleoperate the redundant robot manipulator for the target pose-reaching task, which uses the end-effector's tip to reach the gray dot for each target area while making the orientation of the end-effector as perpendicular as possible to the target area surface. The manipulator begins with the fixed starting configuration. This configuration aligns with the operator’s arm being extended straight forward. The operator is required to sequentially reach each target location in numerical order, following the specified task requirements. The objective of the experiment is to test the accuracy of teleoperation and the generality of the proposed system under different operator anthropometric upper body measurements. Four participants engage in this experiment and have a mean height of $169$±$1.24$ \textit{cm} and a mean arm span of $173.9$±$3.0$ \textit{cm}. Note that participant $1$ engages in training data collection but has not been practiced for the experiment. We hold a preparation phase for each participant, including setting up the Xsens Awinda body tracking system described in~\ref{awinda_data} and measuring each participant's arm and span and full-body height for fine-tuning the body representation profile in the Xsens Awinda software. Then, we provide each participant ten minutes to move their arm to become familiar with the relation between their arm joint angles and the Kinova joint angles. In the real test procedure, each participant is required to finish the task five times. Tasks finishing time, end-effector pose, and right arm joints angle trajectory are recorded for each participant.
Table~\ref{experiment_result} presents a summary of the results. The Euclidean distance quantifies the average distance between the end-effector's tip and the target center position across trials in Cartesian space when it touches the target surface. A smaller Euclidean distance indicates higher positional accuracy. Furthermore, the orientation difference between the target pose and the instant pose when the end-effector's tip touches the target surface is represented using the cosine similarity. A value closer to $1$ indicates that the two orientations are aligned in the same direction. In all trials, participants successfully teleoperated the manipulator to reach the desired target area, with a mean absolute error $2.51$±$0.75$ \textit{cm}, and the mean cosine similarity for all target poses is $0.97$±$0.01$. This supports the generality and accuracy of the proposed system.

\begin{table}[]
\scriptsize 
\begin{tabular}{cllll}
\multirow{2}{*}{\textbf{Participant}} &
  \multirow{2}{*}{\textbf{Time (s)}} &
  \multicolumn{3}{l}{\textbf{\begin{tabular}[c]{@{}l@{}}Euclidean Dist. (cm)/\\ Orientation Diff. (radian)\end{tabular}}} \\ \cline{3-5} 
\noalign{\vskip 2pt}
             &              & \textbf{Target 1} & \textbf{Target 2} & \textbf{Target 3} \\[2pt] \hline
\noalign{\vskip 2pt}
\textbf{1} &
  57.39±1.85 &
  \begin{tabular}[c]{@{}l@{}}1.91±0.39/\\ 0.98±0.00\end{tabular} &
  \begin{tabular}[c]{@{}l@{}}4.09±0.51/\\ 0.97±0.00\end{tabular} &
  \begin{tabular}[c]{@{}l@{}}1.12±0.40/\\ 0.99±0.00\end{tabular} \\[5pt] \hline
\noalign{\vskip 2pt} 
\textbf{2} &
  55.96±5.46 &
  \begin{tabular}[c]{@{}l@{}}2.07±0.29/\\ 0.97±0.01\end{tabular} &
  \begin{tabular}[c]{@{}l@{}}1.50±0.27/\\ 0.96±0.00\end{tabular} &
  \begin{tabular}[c]{@{}l@{}}2.13±0.13/\\ 0.99±0.00\end{tabular} \\[5pt] \hline
\noalign{\vskip 2pt} 
\textbf{3} &
  186.66±5.98 &
  \begin{tabular}[c]{@{}l@{}}2.51±0.75/\\ 0.97±0.01\end{tabular} &
  \begin{tabular}[c]{@{}l@{}}3.53±0.55/\\ 0.99±0.00\end{tabular} &
  \begin{tabular}[c]{@{}l@{}}1.92±0.57/\\ 0.99±0.00\end{tabular} \\[5pt] \hline
\noalign{\vskip 2pt} 
\textbf{4} &
  138.58±27.34 &
  \begin{tabular}[c]{@{}l@{}}2.42±0.34/\\ 0.97±0.00\end{tabular} &
  \begin{tabular}[c]{@{}l@{}}3.67±0.67/\\ 0.98±0.00\end{tabular} &
  \begin{tabular}[c]{@{}l@{}}4.99±1.17/\\ 0.99±0.00\end{tabular} \\[5pt] \hline
\noalign{\vskip 2pt} 
\textbf{All} & 116.17±13.91 & \multicolumn{3}{c}{2.73±0.95 / 0.98±0.00}      
\end{tabular}
\captionsetup{font=footnotesize}
\caption{The standard error of the cosine similarity for orientation reported as $0.00$ is rounding and does not represent an actual zero.}
\label{experiment_result}
\vspace{-8mm}
\end{table}

\vspace{-2.8mm}
\section{limitations}
\label{sec:limitations}
\vspace{-1.9mm}
The initial training for mapping human arm gestures to the manipulator’s latent distribution space showed promising performance for redundant robot manipulator teleoperation. However, the initial training data is collected from only one individual. This introduces potential biases, in other words, different individuals can have distinct conceptions of a robot manipulator joint configuration corresponding to their arm configuration. For example, in the experimental results, participant $3$ took significantly longer to teleoperate the robot manipulator to reach the target pose. We observe that this participant attempted to use a lower back bending motion to control the Joint $0$ of the manipulator, which should ideally correlate with the shoulder joint's abduction/adduction. This variance highlights the need for more diverse data collection across different individuals to capture a broader range of natural human motions to correspond with the robot manipulator joint configuration. A comparable example is the MNIST dataset, where different individuals can interpret handwritten numbers differently, leading to variations in classification outcomes. Incorporating articulated thoracic and pelvic motion as additional features in human gesture mapping to the manipulator’s latent distribution space can be a promising opportunity for future work.
The proposed GRU-based VAE model focuses on approximating the robot's configuration space and generating joint angle trajectories. However, it currently lacks information about joint velocities, which could be crucial for improving the smoothness and responsiveness of the teleoperation. Future work could address this by extending the data collection pipeline to include joint velocity trajectories, as discussed in Section~\ref{kinova_data}. By incorporating velocity information into the latent distribution space, the model could generate smoother and more precise robot manipulator movements, enhancing teleoperation performance.

\vspace{-2.5mm}

\section{conclusion}
\label{sec:conclusion}
\vspace{-1.5mm}
This paper presents a teleoperation solution for a redundant 7-DOF robot manipulator. We developed and implemented a GRU-based VAE architecture finding a latent representation of the manipulator configuration space. The learned latent distribution space can utilize robot combinatorial joint kinematics that can mimic complex human articulated arm motion. In addition, a fully-connected neural network module mapped human arm gestures into latent distribution space representations, thus mimicking and generating corresponding robot manipulator configuration trajectories via the VAE decoder in real-time. The initial training for mapping human arm gestures to the manipulator’s latent distribution space showed promising performance for redundant robot manipulator teleoperation and the decoder was able to generate new robot manipulator configurations based on human operator right arm gestures that were never encountered in the training process.

\vspace{-4.5mm}
\section*{ACKNOWLEDGMENT}
The authors would like to thank all the members of the Center for Distributed Robotics Laboratory for their help. This work is supported by the Minnesota Robotics Institute
(MnRI) and the National Science Foundation through
grants  \#CNS-1531330, \#CNS-1919631, and \#CNS-1939033.
USDA/NIFA has also supported this work through the grants
2020-67021-30755 and 2023-67021-39829.


\clearpage

\bibliographystyle{ieeetr}
\bibliography{kinova_teleop}

\end{document}